\documentclass[sigconf]{acmart}

\AtBeginDocument{%
  \providecommand\BibTeX{{%
    \normalfont B\kern-0.5em{\scshape i\kern-0.25em b}\kern-0.8em\TeX}}}

\copyrightyear{2023} 
\acmYear{2023} 
\setcopyright{acmlicensed}
\acmConference[WWW '23]{Proceedings of the ACM Web Conference 2023}{April 30-May 4, 2023}{Austin, TX, USA}
\acmBooktitle{Proceedings of the ACM Web Conference 2023 (WWW '23), April 30-May 4, 2023, Austin, TX, USA}
\acmPrice{15.00}
\acmDOI{10.1145/3543507.3583412}
\acmISBN{978-1-4503-9416-1/23/04}

\usepackage{newfloat}
\usepackage{listings}
\usepackage{graphicx}
\usepackage{multirow,array}
\usepackage{url}
\usepackage{amsmath}
\usepackage{bbding}
\usepackage{balance}
\usepackage{bm}
\usepackage{bbm}
\usepackage{adjustbox}

\usepackage{enumitem}
\setlist[itemize]{leftmargin=*}

\begin{document}

\title{KRACL: Contrastive Learning with Graph Context Modeling for Sparse Knowledge Graph Completion}



\author{Zhaoxuan Tan}
\affiliation{%
  \institution{Xi'an Jiaotong University}
  \city{Xi'an}
  \country{China}
}
\email{tanzhaoxuan@stu.xjtu.edu.cn}

\author{Zilong Chen}
\affiliation{%
  \institution{Tsinghua University}
  \city{Beijing}
  \country{China}
}
\email{chenzl22@mails.tsinghua.edu.cn}

\author{Shangbin Feng}
\affiliation{%
  \institution{University of Washington}
  \city{Seattle}
  \country{United States}
}
\email{shangbin@cs.washington.edu}

\author{Qingyue Zhang}
\affiliation{%
  \institution{Xi'an Jiaotong University}
  \city{Xi'an}
  \country{China}
}
\email{zhangqingyue2019@stu.xjtu.edu.cn}

\author{Qinghua Zheng}
\affiliation{%
  \institution{Xi'an Jiaotong University}
  \city{Xi'an}
  \country{China}
}
\email{qhzheng@mail.xjtu.edu.cn}

\author{Jundong Li}
\affiliation{%
  \institution{University of Virginia}
  \city{Charlottesville}
  \country{United States}
}
\email{jundong@virginia.edu}

\author{Minnan Luo}
\authornote{Corresponding author: Minnan Luo, School of Computer Science and Technology, Xi’an Jiaotong University, Xi’an 710049, China.}
\affiliation{%
  \institution{Xi'an Jiaotong University}
  \city{Xi'an}
  \country{China}
}
\email{minnluo@xjtu.edu.cn}
\renewcommand{\shortauthors}{Zhaoxuan Tan et al.}

\begin{abstract}
  Knowledge Graph Embeddings (KGE) aim to map entities and relations to low dimensional spaces and have become the \textit{de-facto} standard for knowledge graph completion. Most existing KGE methods suffer from the sparsity challenge, where it is harder to predict entities that appear less frequently in knowledge graphs.
In this work, we propose a novel framework KRACL\footnote{The code is available at \url{https://github.com/TamSiuhin/KRACL}.} to alleviate the widespread sparsity in KGs with graph context and contrastive learning. Firstly, we propose the Knowledge Relational Attention Network (KRAT) to leverage the graph context by simultaneously projecting neighboring triples to different latent spaces and jointly aggregating messages with the attention mechanism. KRAT is capable of capturing the subtle semantic information and importance of different context triples as well as leveraging multi-hop information in knowledge graphs. 
Secondly, we propose the knowledge contrastive loss by combining the contrastive loss with cross entropy loss, which introduces more negative samples and thus enriches the feedback to sparse entities. 
Our experiments demonstrate that KRACL achieves superior results across various standard knowledge graph benchmarks, especially on WN18RR and NELL-995 which have large numbers of low in-degree entities. Extensive experiments also bear out KRACL's effectiveness in handling sparse knowledge graphs and robustness against noisy triples.
\end{abstract}




\maketitle

\section{Introduction}
Knowledge graphs (KGs) are collections of large-scale facts in the form of structural triples \textit{(subject, relation, object)}, denoted as $(s, r, o)$, \emph{e.g.}, \textit{(Christopher Nolan, Born-in, London)}. These KGs reveal the relations between entities and play an important role in many applications such as natural language processing~\cite{zhou2022eventbert,ren2021learning,kgbart,KCD}, computer vision~\cite{vision,vision2}, and recommender systems~\cite{wang2020reinforced, xia2021knowledge, geng2022pathrecomm, wang2019multi,cao2019unifying}.
Although KGs already contain millions of facts, they are still far from complete, \emph{e.g.}, 71\% of people in the Freebase knowledge graph have no birthplace and 75\% have no nationality \cite{FBdong}, which leads to poor performance on downstream applications. Therefore, knowledge graph completion (KGC) is an important task to predict whether a given triple is valid or not and further expands the existing KGs. \par

Most existing KGs are stored in symbolic form while downstream applications always involve numerical computation in continuous spaces. To address this issue, researchers proposed to map entities and relations to low dimensional embeddings dubbed knowledge graph embedding (KGE). These models usually leverage geometric properties in latent space, such as translation and bilinear transformation in Euclidean space \cite{transe, distmult}, rotation in complex space \cite{rotate}, and reflection in hypersphere space \cite{low_hyperbolic}. Convolutional networks are also used in KGE to extract semantic information \cite{conve, convkb}.
Recently, Graph Neural Networks (GNNs) are leveraged to encode graph structure in KGE \cite{compgcn, zhang2020relational} and propagate message in path-based knowledge graph reasoning \cite{zhu2021neural, zhang2022knowledge}.




\begin{figure}[t]
    \centering
    \includegraphics[width=1\linewidth]{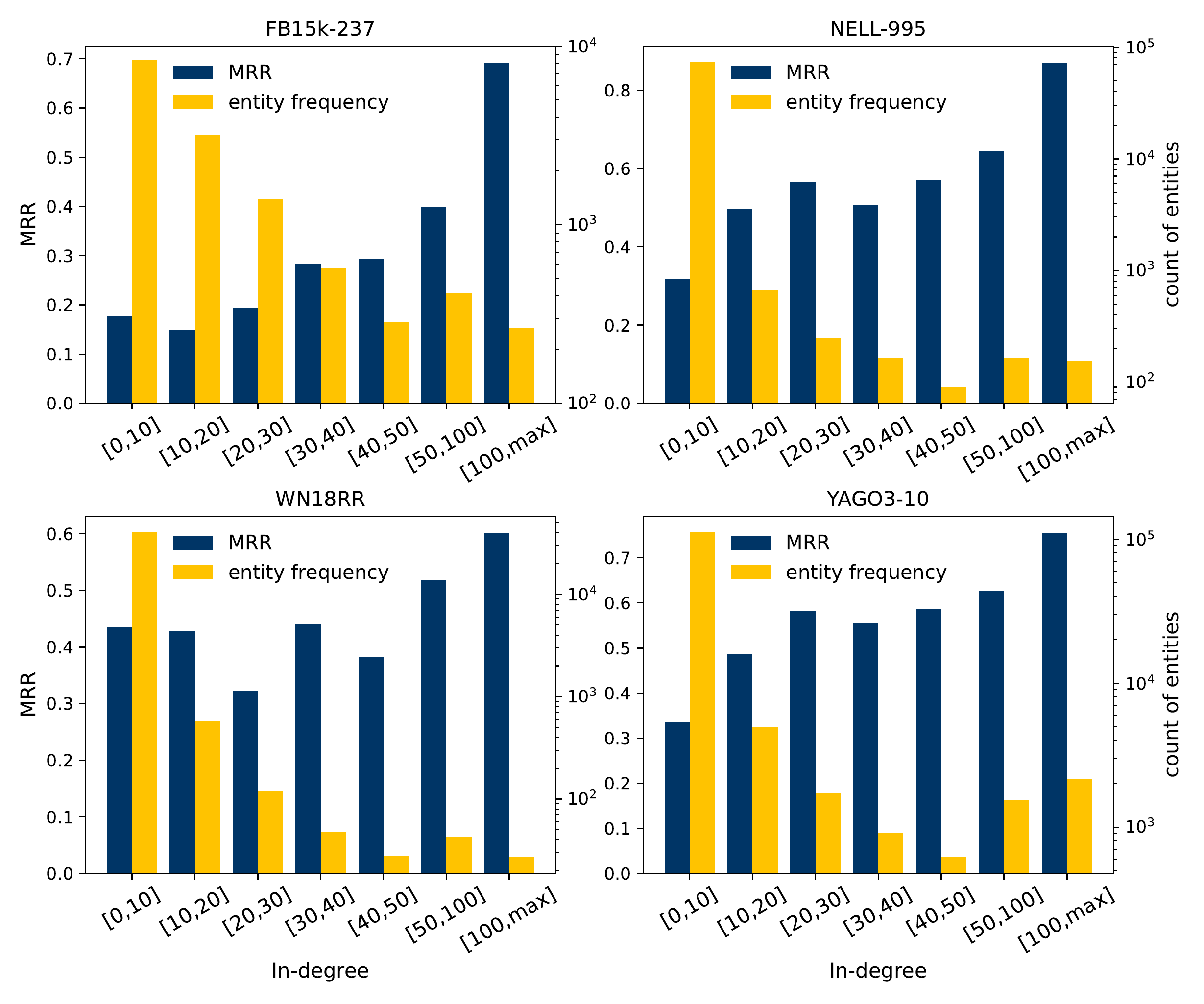}
    \caption{The number and mean reciprocal rank (MRR) of different frequency entities based on RotatE results on FB15k-237, WN18RR, NELL-995, and YAGO3-10 benchmark datasets. It reveals the common existence of sparse entities and their poor prediction performance in KGs.}
    \label{fig:teaser}
\end{figure}

Although much research progress has been made by recent KGE models, predicting entities that rarely appear in knowledge graphs remains challenging \cite{whereKGEfall}. We investigate the in-degree (using entity frequency) and link prediction performance (using MRR) on several widely acknowledged knowledge graphs, including FB15k-237 \cite{fb15k237}, WN18RR \cite{conve}, NELL-995 \cite{nell995}, and YAGO3-10 \cite{YAGO310} (shown in Figure \ref{fig:teaser}). The yellow bars show that a large portion of entities rarely appear in knowledge graph triples, leading to limited facts for knowledge graph completion. Moreover, it also reveals the common existence of sparse entities across various datasets. The blue bars show the link prediction performance for entities of different in-degree with RotatE \cite{rotate}. 
We observe that the prediction performance is strongly relevant to the entity in-degree, and the prediction performance of sparse entities is much worse than those of frequent entities.

In this work, we propose \textbf{KRACL} (\textbf{K}nowledge \textbf{R}elational \textbf{A}ttention Network with \textbf{C}ontrastive \textbf{L}earning) to alleviate the sparsity issue in KGs. First, we employ \textbf{K}nowledge \textbf{R}elational \textbf{AT}tention Network (KRAT) to fully leverage the graph context in KG. Specifically, we map context triples to different latent spaces and combine them in message, then we calculate attention score for each context triple to capture its importance and aggregate messages with the attention scores to enrich the sparse entities' embedding.
Second, we project subject entity embedding to object embedding with knowledge projection head, \emph{e.g.}, ConvE, RotatE, DistMult, and TransE. Finally, we optimize the model with proposed knowledge contrastive loss, \emph{i.e.}, combining the contrastive loss and cross entropy loss. We empirically find that contrastive loss can provide more feedback to sparse entities and is more robust against sparsity when compared to explicit negative sampling. Extensive experiments on various standard benchmarks show the superiority of our proposed KRACL model over competitive peer models, especially on WN18RR and NELL-995 with many low in-degree nodes. Our key contributions are summarized as follows:

\begin{itemize}
    \item We propose the \textbf{K}nowledge \textbf{R}elational \textbf{AT}tention Network (KRAT) to integrate knowledge graph context by mapping neighboring triples to different representation space, combining different latent spaces in message, and aggregating message with the attention mechanism. To the best of our knowledge, we are the first to ensemble different KG operators in the GNN architecture.
    
    \item We propose a knowledge contrastive loss to alleviate the sparsity of knowledge graphs. We incorporate contrastive loss with cross entropy loss to introduce more negative samples, which can enrich the feedback to limited positive triples in knowledge graphs, thus enhancing prediction performance for sparse entities.
    
    \item Experimental results demonstrate that our proposed KRACL framework achieves superior performance on five standard benchmarks, especially on WN18RR and NELL-995 with many low in-degree entities.
\end{itemize}

\begin{figure*}[t]
    \centering
    \includegraphics[width=1\linewidth]{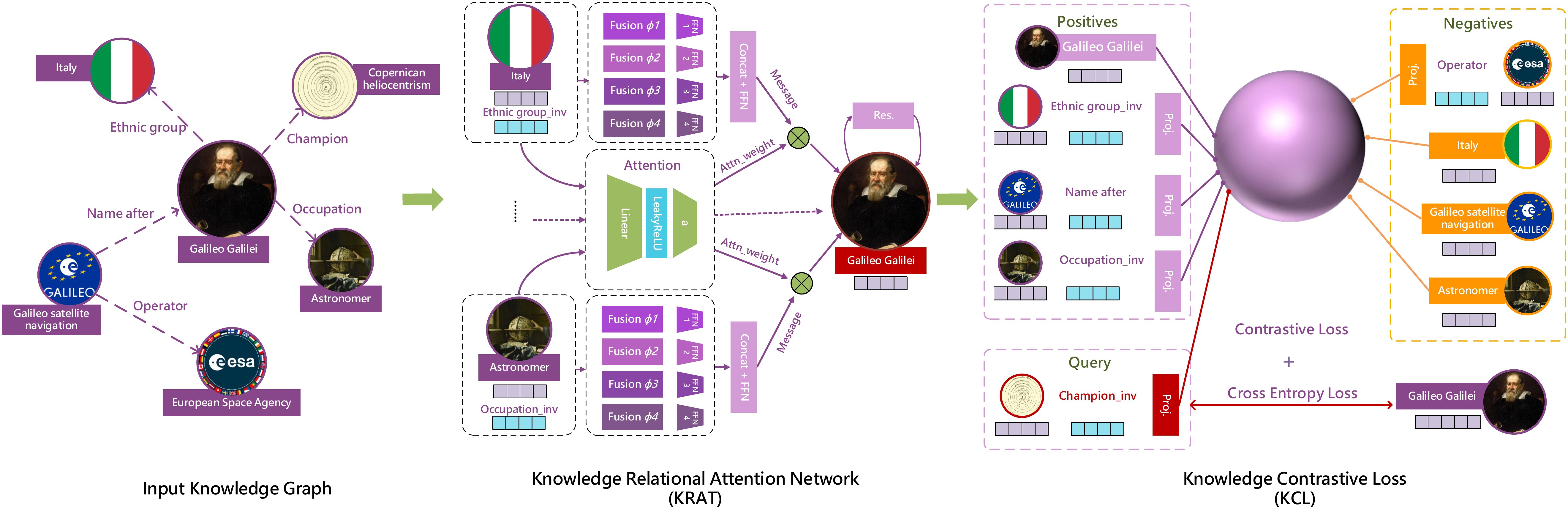}
    \caption{Overview of our proposed KRACL framework to alleviate the sparsity problem in knowledge graphs.}
    \label{fig:my_label}
\end{figure*}

\section{Related Work}
\subsection{Knowledge Graph Embedding}
\noindent\textbf{Non-Neural}
Non-neural models embed entities and relations into latent space with linear operations. Starting from TransE \cite{transe}, the pioneering and most representative translational model, a series of models are proposed in this line, such as TransH \cite{transh}, and TransR \cite{transr}. RotatE \cite{rotate} extends the translational model to complex space and OTE \cite{OTE} further extends RotatE to high dimensional space. There is another line of work that takes tensor decomposition to compute the plausibility of triples. For instance, RESCAL \cite{rescal} and DistMult \cite{distmult} represent each relation with a full rank matrix and diagonal matrix, respectively. ComplEx \cite{complex} generalizes DistMult to complex space to enhance the expressiveness. To model the uncertainty of learned representations, Gaussian distribution is also used for KGE \cite{gaussian}.
Beyond Euclidean space and Gaussian distribution, KGE can also be projected to manifold \cite{manifoldembedding}, Lie group \cite{transg}, hyperbolic space \cite{murp, low_hyperbolic}, and mixture of different embedding spaces \cite{GIE}. These models are simple and intuitive, but due to their simple operation and limited parameters, these non-neural models usually produce low-quality embeddings.
%

\noindent\textbf{Neural Network-based} Neural network-based KGE models are introduced for KGC due to their inherent strong learning ability. 
Convolutional neural networks are employed to extract the semantic features from KGE. Specifically, ConvE \cite{conve} utilizes 2D convolution to learn deep features of entities and relations. ConvKB \cite{convkb} adopts 1D convolution and feeds the whole triple into the convolutional neural network. HypER \cite{hyper} employs hypernetwork to generate relation-special filters.
Graph neural networks also show strong potential in learning knowledge graphs embedding by incorporating graph structure in KGs \cite{sacn, a2n, DisenKGAT}.
R-GCN \cite{rgcn} is an extension of the graph convolution network \cite{gcn} for relational data. 
CompGCN \cite{compgcn} jointly embeds both entities and relations in KG through compositional operators. 
NBFNet \cite{zhu2021neural} and RED-GNN \cite{zhang2022knowledge} leverage GNNs to progressively propagate query in knowledge graph to conduct KG reasoning.
In light of recent advance, capsule networks \cite{vu2019capsule}, transformers \cite{nayyeri20215, chen2020hitter}, language models \cite{yao2019kgbert, kgt5}, and commonsense \cite{cake} are also introduced to KGC. 

\subsection{Contrastive Learning}
Contrastive learning has been a popular approach for self-supervised learning by pulling semantically close neighbors together while pushing apart non-neighbors away \cite{contrastivehadsell}. As is first introduced in the computer vision domain, a large collection of works \cite{mocov1, simclr, contrastivemulti} learn self-supervised image representations by minimizing the distance between two augmented views of the same image, while pushing apart the two augmented views of the same images. \citet{supervisedCL} further extends contrastive learning to the supervised setting by considering the representations from the same class as positive samples. Contrastive learning also achieves great success in different fields, including natural language processing \cite{simcse, clnlp3, xu2022sequence}, graph representation learning \cite{gcl,gca,yang2022dual,xia2022simgrace,park2022cgc,li2022graph,wang2022clusterscl}, multi-modal pre-training \cite{radford2021learning}, and stance detection \cite{liang2022zero}. 
In the knowledge graph area, contrastive learning has been applied for recommendation \cite{yang2022knowledge}, molecular representation learning \cite{fang2022molecular}, text-based KGC \cite{wang2022simkgc}, and efficient KGE \cite{wang2022swift}. We explore its potential to alleviate knowledge graphs' sparsity in this work.

\section{Methodology}
We consider a knowledge graph as a collection of factual triples $\mathcal{D}=\{(s,r,o)\}$ with $\mathcal{E}$ as entity set and $\mathcal{R}$ as relation set. Each triple has a subject entity $s$ and object entity $o$, where $s,o\in \mathcal{E}$. Relation $r\in \mathcal{R}$ connects two entities with direction from subject to object. 
Next, we introduce a novel framework--\textbf{K}nowledge \textbf{R}elational \textbf{A}ttention Network with \textbf{C}ontrastive \textbf{L}earning (KRACL) for knowledge graph completion. 
KRACL is two-fold, we first introduce the \textbf{K}nowledge \textbf{R}elational \textbf{AT}tention Network (KRAT) that aggregates the graph context information in KG, then we describe \textbf{K}nowledge \textbf{C}ontrastive \textbf{L}earning (KCL) to alleviate the sparsity problem.

\subsection{Knowledge Relational Attention Network}
To fully exploit the limited context information in sparse KGs, we first use different operators to project the context triples into different representation spaces and merge their inductive bias, then combine them in the message. Then we aggregate all the context triples with the attention mechanism. The entities' KRAT update equation can be defined as
\begin{equation}
    \boldsymbol{h}_o^{(l)} = \sigma\left(\sum_{(s, r)\in \mathcal{N}_o}\alpha_{sro}\boldsymbol{W}_{agg}^{(l)}f(\boldsymbol{h}_s, \boldsymbol{h}_r) + \boldsymbol{W}_{res}^{(l)}\boldsymbol{h}_o^{(l-1)} \right),
    \label{aggregation}
\end{equation}
where $\sigma$ denotes the Tanh activation function, $\boldsymbol{W}_{agg}^{(l)}\in \mathbbm{R}^{d_{l}\times n\cdot d_{l-1}}$ denotes the feed-forward aggregation matrix, $\alpha_{sro}$ denotes the attention weights of context triple $(s,r,o)$. We also add a pre-activation residual connection to prevent over-smoothing. The aggregated message $f(\boldsymbol{h}_s, \boldsymbol{h}_r)$ that combines KG context aims to map neighboring triples to different latent spaces with fusion operators and fully leverage their semantic information, which can be written as
\begin{equation}
    f(\boldsymbol{h}_s, \boldsymbol{h}_r) = \sigma\left(\left[\boldsymbol{W}_1^{(l)}\phi_{1}(\boldsymbol{h}_s, \boldsymbol{h}_r)\big|\big|...\big|\big|\boldsymbol{W}_{n}^{(l)}\phi_{n}(\boldsymbol{h}_s, \boldsymbol{h}_r)\right]\right),
\end{equation}
where $\sigma$ denotes the LeakyReLU activation function, $\boldsymbol{W}_i^{(l)}\in \mathbbm{R}^{d_{l}\times d_{l-1}}$ denotes the weighted matrix corresponding to the $i$-th operator in the $l$-th layer, $\phi_{i}(\boldsymbol{h}_s, \boldsymbol{h}_r)$ denotes the fusion operation between subject $s$ and relation $r$, Then we concatenate them and merge their inductive bias to serve as the passing message in KRAT layer. The composition operator $\phi_{i}(\boldsymbol{h}_s, \boldsymbol{h}_r)$ can be subtraction taken from \citet{transe}, multiplication taken from \citet{distmult}, rotation taken from \citet{rotate}, and circular-correlation taken from \citet{hole}\footnote{See details of rotation and circular-correlation operations in Appendix.}:

\begin{itemize}
    \item Subtraction (Sub): $\phi(\boldsymbol{h}_s, \boldsymbol{h_r}) = \boldsymbol{h}_s - \boldsymbol{h}_r$
    \item Multiplication (Mult): $\phi(\boldsymbol{h}_s, \boldsymbol{h_r}) = \boldsymbol{h}_s \cdot \boldsymbol{h}_r$
    \item Rotation (Rot): $\phi(\boldsymbol{h}_s, \boldsymbol{h_r}) = \boldsymbol{h}_s \circ \boldsymbol{h}_r$
    \item Circular-correlation (Corr): $\phi(\boldsymbol{h}_s, \boldsymbol{h_r}) = \boldsymbol{h}_s \star \boldsymbol{h}_r$
\end{itemize}


 We then calculate the attention score for each context triple $\alpha_{sro}$ to distinguish their importance. Inspired by \citet{gat, KBAT}, the attention score $w_{sro}$ for context triple $(s,r,o)$ is defined as
\begin{equation}
    w_{sro} = \bm{a}^{(l)}  LeakyReLU\left(\bm{W}_{att}^{(l)}\left[\bm{h}_s^{(l-1)}||\bm{h}_r^{(l-1)}||\boldsymbol{h}_o^{(l-1)}\right]\right),
\end{equation}
where $\boldsymbol{a}^{(l)}\in\mathbbm{R}^{1\times d_e}$ and $\boldsymbol{W}_{att}^{(l)}\in \mathbbm{R}^{d_e\times (2d_e+d_r)}$ are learnable parameters specific for the $l$-th layer of KRAT, $\boldsymbol{h}_s,\boldsymbol{h}_r, \boldsymbol{h}_o$ denote the hidden representations of subject entity, relation, and object entity in the $l-1$ layer. Then the attention score of each triple is normalized with softmax as
\begin{equation}
    \begin{split}
        \alpha_{sro} & = softmax_{sr}(w_{sro})\\
    & =\frac{exp(w_{sro})}{\sum_{n\in \mathcal{N}_o}\sum_{p\in \mathcal{R}_{no}} exp(w_{npo})},
    \end{split}
\end{equation}
where $\mathcal{N}_o$ denotes the neighbor entities of $o$, $\mathcal{R}_{no}$ denotes the relation that connects with $n$ and $o$, $\alpha_{sro}$ is the normalized attention weight for triple $(s,r,o)$. 


After the entity embedding update defined in Eq \ref{aggregation}, the relation representations are also transformed as follows:
\begin{equation}
    \boldsymbol{h}_r^{(l)} = \boldsymbol{W}_{rel}^{(l)}\cdot \boldsymbol{h}_r^{(l-1)},
\end{equation}
where $\boldsymbol{W}_{rel}^{(l)}\in \mathbbm{R}^{d_{r}\times d_{r}}$ is a trainable weight matrix for relation embeddings under the $l$-th layer.

\subsection{Knowledge Contrastive Learning}
After passing $T$ layers of KRAT, the entity representations are enriched with $T$ hops context. Taking the idea of supervised contrastive learning \cite{supervisedCL}, we pulls embeddings from the same entities close and pushes embeddings from different entity further away to introduce more negative samples as well as the feedback to limited positive triples in KG. The contrastive loss is calculated as 
\begin{equation}
    \mathcal{L}_{CL}=\sum_{o\in \mathcal{T}}\frac{-1}{|\mathcal{T}_o|}\sum_{\boldsymbol{z}_{(s,r)}\in \mathcal{T}_o} log\frac{exp(\boldsymbol{z}_{(s,r)}\cdot \boldsymbol{h}_o/\tau)}{\sum\limits_{k\notin \mathcal{T}_o}{exp(\boldsymbol{z}_k\cdot \boldsymbol{h}_o/\tau)}},
    \label{SCL}
\end{equation}
where $\mathcal{T}$ denotes a batch of normalized entity embeddings, $\mathcal{T}_o$ denotes the set of representations corresponding to entity $o$, $\tau$ is an adjustable temperature hyperparameter that controls the balance between uniformity and tolerance \cite{understandcl}. The contrastive loss introduces more negative samples, therefore enriching the feedback to the limited positive triples.
$z_{(s,r)}$ is a knowledge projection head which can be TransE, DistMult, RotatE, and ConvE to transform embeddings from subject to object. Here we take ConvE as an example, 
\begin{equation}
    \boldsymbol{z}_{(s, r)} = \sigma(vec(\sigma( \left[ \overline{\boldsymbol{h}_s} \big| \big| \overline{\boldsymbol{h}_r} \right]\ast\omega))\boldsymbol{W}_p),
\end{equation}
where $\overline{\boldsymbol{h}_s}\in \mathbbm{R}^{d_w\times d_h}$ and $\overline{\boldsymbol{h}_r}\in \mathbbm{R}^{d_w\times d_h}$ denote 2D reshaping of $\boldsymbol{h}_s\in \mathbbm{R}^{d_w d_h\times 1}$ and $\boldsymbol{h}_r\in \mathbbm{R}^{d_w d_h\times 1}$ respectively, $[\cdot || \cdot]$ is concatenation operation, $\ast$ denotes the convolution operation, $\sigma$ denotes non-linearity (PReLU \cite{prelu} by default), $vec$ denotes vectorization, and $\boldsymbol{W}_p$ is a linear transformation matrix. The whole formula represents the predicted object representation given the subject $s$ and relation $r$. 
 We then calculate the cross entropy loss as follows
\begin{equation}
    \mathcal{L}_{CE} = -\frac{1}{|\mathcal{T}|}\sum_{(s,r)\in \mathcal{T}}\sum_{o\in \mathcal{E}}y_{(s,r)}^{o}\cdot \log \hat{y}_{(s,r)}^{o},
\end{equation}
where $\mathcal{T}$ denotes training triples in a batch, $\mathcal{E}$ denotes all entities that exist in the KG, $y_{(s,r)}^{o}$ denotes the ground-truth labels, \emph{i.e.}, $y_{(s,r)}^{o}=1$ if triple $(s,r,o)$ is valid and $y_{(s,r)}^{o}=0$ otherwise. $z_{(s,r)}$ is 1-N scoring function taken from ConvE \cite{conve}, which scores all candidate entities with dot product. It is also used for scoring in inference
\begin{equation}
    \hat{y}_{(s,r)}^o = \boldsymbol{z}_{(s,r)}\cdot \boldsymbol{h}_o^T,
\end{equation}
where $\hat{y}_{(s,r)}^o$ denotes the predicted plausibility for triple $(s,r,o)$, $\boldsymbol{h}_o\in \mathbbm{R}^{d\times |\mathcal{E}|}$ denotes the representations of all entities. Ultimately, we demonstrate the final objective by incorporating the contrastive loss and cross entropy loss through summation,
\begin{equation}
    \mathcal{L} = \mathcal{L}_{CL} + \mathcal{L}_{CE}.
\end{equation}
By jointly optimizing the two objectives, we capture the similarity of the embeddings corresponding to the same entity and contrast them with other entities, while boosting the performance for link prediction.

\begin{table}[t]
    \centering
    \caption{Dataset statistics.}
    \begin{adjustbox}{max width=0.48\textwidth}
    \begin{tabular}{l c c c c c c c}
         \toprule[1.5pt]
         \multirow{2}{*}{\textbf{Dataset}} & \multirow{2}{*}{\textbf{\#Ent.}} & \multirow{2}{*}{\textbf{\#Rel.}} & \multicolumn{3}{c}{\textbf{\#Edge}} & 
         \multicolumn{2}{c}{\textbf{\#In-degree}}\\
         \cmidrule(r){4-6} \cmidrule(r){7-8}
         & & & Train & Valid & Test & Avg. & Med. \\
         \midrule[0.75pt]
         \textbf{FB15k-237}& 14,541 & 237 & 272,115 & 17,535 & 20,466 & 18.76 & 8 \\
         \textbf{WN18RR} & 40,943 & 11 & 86,835 & 3,034 & 3,134 & 2.14 & 1 \\
         \textbf{NELL-995} & 75,492 & 200 & 149,678 & 543 & 3,992 & 2.01 & 0 \\
         \textbf{Kinship} & 104 & 25 & 8,544 & 1,068 & 1,074 & 82.15 & 82\\
         \textbf{UMLS} & 135 & 46 & 5,216 & 652 & 661 & 38.63 & 20\\
         \bottomrule[1.5pt]
    \end{tabular}
    \end{adjustbox}
    
    \label{tab:benchmark}
    
\end{table}

\begin{table*}[t]
    \centering
     \caption{Knowledge graph completion performance on sparse knowledge graphs, \emph{i.e.}, WN18RR and NELL-995. The best score is in \textbf{bold} and the second best score is \underline{underlined}, `-' indicates the result is not reported in previous work.}
    \begin{tabular}{l|c c c c c|c c c c c}
         \toprule[1.5pt]
         \multirow{2}{*}{\textbf{Model}} & \multicolumn{5}{c|}{\textbf{WN18RR}} & \multicolumn{5}{c}{\textbf{NELL-995}} \\
         & MRR $\uparrow$ & MR $\downarrow$& H@10 $\uparrow$ & H@3 $\uparrow$ & H@1 $\uparrow$ & MRR $\uparrow$ & MR $\downarrow$ & H@10 $\uparrow$ & H@3 $\uparrow$ & H@1 $\uparrow$ \\ 
         \midrule[0.75pt]
          TransE \citep{transe}& .243 & 2300 & .532 & .441 & .043 & .401 & 2100 & .501 & .472 & .344 \\
          DistMult \citep{distmult}& .444 & 7000 & .504 & .470 & .412  & .485 & 4213 & .610 & .524 & .401\\
          ComplEx  \citep{complex}& .449 & 7882 & .530 & .469 & .409 &  .482 & 4600 & .606 & .528 & .399\\
         RotatE \citep{rotate}& .494 & 4046 & .571 & .510 & .455 & .483 & 2582 &.565 &.514 & .435\\
          ConvE \citep{conve}& .456 & 4464 & .531 & .470 & .419 & .491 & 3560 & .613 & .531 & .403\\
          HypER \citep{hyper}& .493 & 4687 & .549 & .503 & \underline{.464} & .540 & 1763 & .657 & .580 & .471\\
          TuckER \citep{tucker}& .470 & - & .526 & .482 & .443 & .520 & 2330 & .624 & .561& .455\\
         R-GCN \citep{rgcn}& .123 & 6700 & .207 & .137 & .08 & .12 & 7600 & .188 & .126 & .082 \\
          KBAT \citep{KBAT}&.412 & \underline{1921}&  .554 & - & - & .319 & 3683 & .474 & .370 & .233\\
          CompGCN \citep{compgcn}& .481 & 3113 & .548 & .492 & .448 & .534 & 1246 & .644 & \textbf{.607} & .466\\
          HAKE \citep{HAKE}& .497 & -& .582 & .516 & .452 &.508 & 5836& .613& .557&.442 \\
          GC-OTE \citep{OTE}& .491 & - & .583 & .511 & .442 & .538 & 837 & .657 & .576 & .469\\
          HittER \citep{chen2020hitter}& .503 & - & .584 & .516 & .462 & - & - & - & - & - \\
          DisenKGAT \citep{DisenKGAT}& \underline{.506} & 4135 & \underline{.590} & \underline{.522} & .462 & \underline{.547} & 882& \underline{.666} & .598 & .474\\
          GIE \citep{GIE}& .491 & - & .575 & .505 & .452 & .474 & 2218 & .596 & .504 & .408 \\
          CAKE \citep{cake}& - & - & - & - & - & .543 & \textbf{433} & .655 & .583 & \underline{.477} \\
          \hline
          \textbf{KRACL (Ours)} & \textbf{.527} & \textbf{1388} & \textbf{.613} & \textbf{.547} & \textbf{.482} &\textbf{.563} & \underline{716} & \textbf{.672} & \underline{.602} & \textbf{.495}\\
         \bottomrule[1.5pt]
    \end{tabular}
    \label{tab:lp_results1}
\end{table*}

\section{Experiments}

We provide empirical results to demonstrate the effectiveness of our proposed KRACL model. The experiments are designed to answer the following research questions:
\begin{itemize}
    \item \textbf{RQ1:} How does KRACL perform on sparse knowledge graphs, compared to the state-of-the-art KGE models?
    \item \textbf{RQ2:} What is KRACL's impact on the sparse entities in knowledge graphs? 
    \item \textbf{RQ3:} Is KRACL robust to the sparsity, noisy triples, and the choice of knowledge projection head in knowledge graphs? 
\end{itemize}

\begin{table*}[t]
    \centering
    \caption{Knowledge graph completion performance on denser knowledge graphs, \emph{i.e.}, FB15k-237 and Kinship. The best score is in \textbf{bold} and the second best score is \underline{underlined}. `-' indicates the result is not reported in previous work.}
    \begin{tabular}{l|c c c c c| c c c c c}
         \toprule[1.5pt]
         \multirow{2}{*}{\textbf{Model}} &  \multicolumn{5}{c|}{\textbf{FB15k-237}} & \multicolumn{5}{c}{\textbf{Kinship}}\\
         & MRR $\uparrow$ & MR $\downarrow$ & H@10 $\uparrow$& H@3 $\uparrow$& H@1 $\uparrow$& MRR $\uparrow$& MR $\downarrow$ & H@10 $\uparrow$& H@3 $\uparrow$& H@1 $\uparrow$\\ 
         \midrule[0.75pt]
          TransE \citep{transe} & .294 & 357 & .465 & - & - & .211 & 38.9  & .470 & .252 & .093\\
          DistMult \citep{distmult} & .241 & 254 & .419& .263 & .155& .48 & 7.9 & .708 & .491 & .377\\
          ComplEx \citep{complex}& .247 & 339 & .428 &.275 & .158 & .823 & 2.48 & .971 & .899 & .733\\
          RotatE \citep{rotate}& .338 & 177 & .533 & .375 & .241& .738 & 2.9 & .954 & .827 & .617\\
          ConvE \citep{conve}&.325& 244& .501 &.356 &.237 & .772 & 3.0 & .950 & .858 & .665 \\
          HypER \citep{hyper}& .341 & 250 & .520 &.376 &.252 & .868 & 1.96 & .981 & .935 & .790\\
         TuckER \citep{tucker}& .355 & \underline{152} & .541 & .390 & .262& \underline{.885} & \underline{1.67} & \underline{.986} & \underline{.948} & \underline{.816}\\
          R-GCN \citep{rgcn}& .248 & 339 & .428 & .275 & .158 & .109 & 25.9 & .239 & .088 & .030\\
          KBAT \citep{KBAT}& .156 & 392 & .305 & .167 & .085 & .637 & 3.41 & .955 & .757 & .470\\
          CompGCN \citep{compgcn}& .355 & 197 & .535 &.390 & .264 & .810 & 
          2.26 & .977 & .892 & .709\\
          HAKE \citep{HAKE}& .346& - & .542 & .381 & .250 & .802 &2.38 & .968&.881 &.704 \\
          GC-OTE \citep{OTE}& .361 & - & .550 & .396 & .267 & .832& 2.05& .984& .917& .735 \\
           HittER \citep{chen2020hitter}& \textbf{.373} & - & \textbf{.558} & \textbf{.409} & \textbf{.279} & - & - & - & - & -\\
           DisenKGAT \citep{DisenKGAT}& \underline{.368} & 179 & \underline{.553} & \underline{.407} & \underline{.275} & .832 & 1.96 & \underline{.986} & .914 & .737\\
          GIE \citep{GIE}& .362 & - & .552 & .401 & .271 & .664 & 3.43 & .927 & .770 & .520 \\
          CAKE \citep{cake}& .321 & 170 & .515 & .355 & .226 &- &- & -& -& - \\
          \hline
          \textbf{KRACL (Ours)}& .360 & \textbf{150} & .548 & .395 & .266 & \textbf{.895} & \textbf{1.48} &\textbf{.991} &\textbf{.970} & \textbf{.817}\\
         \bottomrule[1.5pt]
    \end{tabular}
    \label{tab:lp_results2}
\end{table*}

\subsection{Experiment Settings}
\subsubsection{\textbf{Datasets.}} To evaluate our KRACL, we consider five widely acknowledged datasets: FB15k-237 \cite{fb15k237}, WN18RR \cite{conve}, NELL-995 \cite{nell995}, Kinship \cite{kinship}, and UMLS \cite{UMLS}, following the standard train/test split. Statistics of these benchmarks are listed in Table \ref{tab:benchmark}, we further investigate the average and medium entity in-degree to demonstrate their sparsity. It is shown that WN18RR and NELL-995 are much sparser than FB15k-237, Kinship, and UMLS.


\subsubsection{\textbf{Evaluation Protocol.}} 
Following \citet{transe}, we use the filtered setting for link prediction, \emph{i.e.}, while evaluating test triples, all valid triples are filtered out from the candidate set. We report mean reciprocal rank (MRR), mean rank (MR), and Hits@N. MRR is the average inverse of obtained ranks of correct entities among all candidate entities. MR means the average obtained ranks of correct entities among all candidate entities. Hits@$N$ measures the proportion of correct entities ranked in the top $N$ among all candidate entities. We take $N$=1,3,10 in this work. 

\subsubsection{\textbf{Baselines.}}
We compare our KRACL with state-of-the-art KGE models, including non-neural model TransE \cite{transe}, DistMult \cite{distmult}, ComplEx \cite{complex}, RotatE \cite{rotate}, TuckER \cite{tucker}, OTE \cite{OTE}, HAKE \cite{HAKE}, and GIE \cite{GIE}; neural network-based model ConvE \cite{conve}, HypER \cite{hyper}, R-GCN \cite{rgcn}, KBAT \cite{KBAT}, CompGCN \cite{compgcn}, HittER \cite{chen2020hitter}, DisenKGAT \cite{DisenKGAT}, and CAKE \cite{cake}.

\begin{itemize}
    \item\textbf{TransE} \cite{transe} is the most representative KGE model with the assumption that the superposition of head and relation embedding is close to tail embedding.
    \item\textbf{DistMult} \cite{distmult} is a matrix factorization model that uses a bilinear function for scoring.
    \item\textbf{ComplEx} \cite{complex} is a matrix factorization model that extends DistMult to the complex space.
    \item\textbf{RotatE} \cite{rotate} is a translational model that maps relations embedding as rotation operation in complex space.
    \item\textbf{ConvE} \cite{conve} is a CNN-based model that adopts 2D convolution network to extract semantic information between entities and relations.
    \item\textbf{HypER} \cite{hyper} is a CNN-based model that uses hypernetwork to construct relational convolution kernels.
    \item\textbf{TuckER} \cite{tucker} is a tensor decomposition model based on TuckER decomposition of binary representation of KG triples.
    \item\textbf{R-GCN} \cite{rgcn} is a GNN-based model that extends GCN to relational data. Specifically, it aggregates messages from different relations with different projection matrices.
    \item\textbf{KBAT} \cite{KBAT} is a GNN-based model that introduces the attention mechanism to learn the importance of neighboring nodes and takes advantage of multi-hop neighbors. Unfortunately, it was found to have test data leakage problem \cite{reeval}.
    \item\textbf{CompGCN} \cite{compgcn} is a GNN-based model that jointly aggregates entity and relation embeddings and score triples with a decoder such as TransE, DistMult, and ConvE.
    \item\textbf{HAKE} \cite{HAKE} maps entities into the polar coordinate system. The radial coordinate aims to model hierarchy and the angular coordinate aims to distinguish entities within the same hierarchy.
    \item\textbf{HittER} \cite{chen2020hitter} is a transformer-based KGE model that is organized in a hierarchical fashion. 
    \item\textbf{DisenKGAT} \cite{DisenKGAT} is a GNN-based model that proposes to leverage micro-disentanglement and macro-disentanglement for representative embeddings.
    \item\textbf{GIE} \cite{GIE} is a translational model that better learns the spatial structures interactively between Euclidean, hyperbolic, and hyperspherical spaces.
    \item\textbf{CAKE} \cite{cake} is a framework that extracts commonsense from factual triples with entities concepts. It can augment negative sampling and joint commonsense and fact-view link prediction.
\end{itemize}


\begin{figure}[t]
    \centering
    \includegraphics[width=1\linewidth]{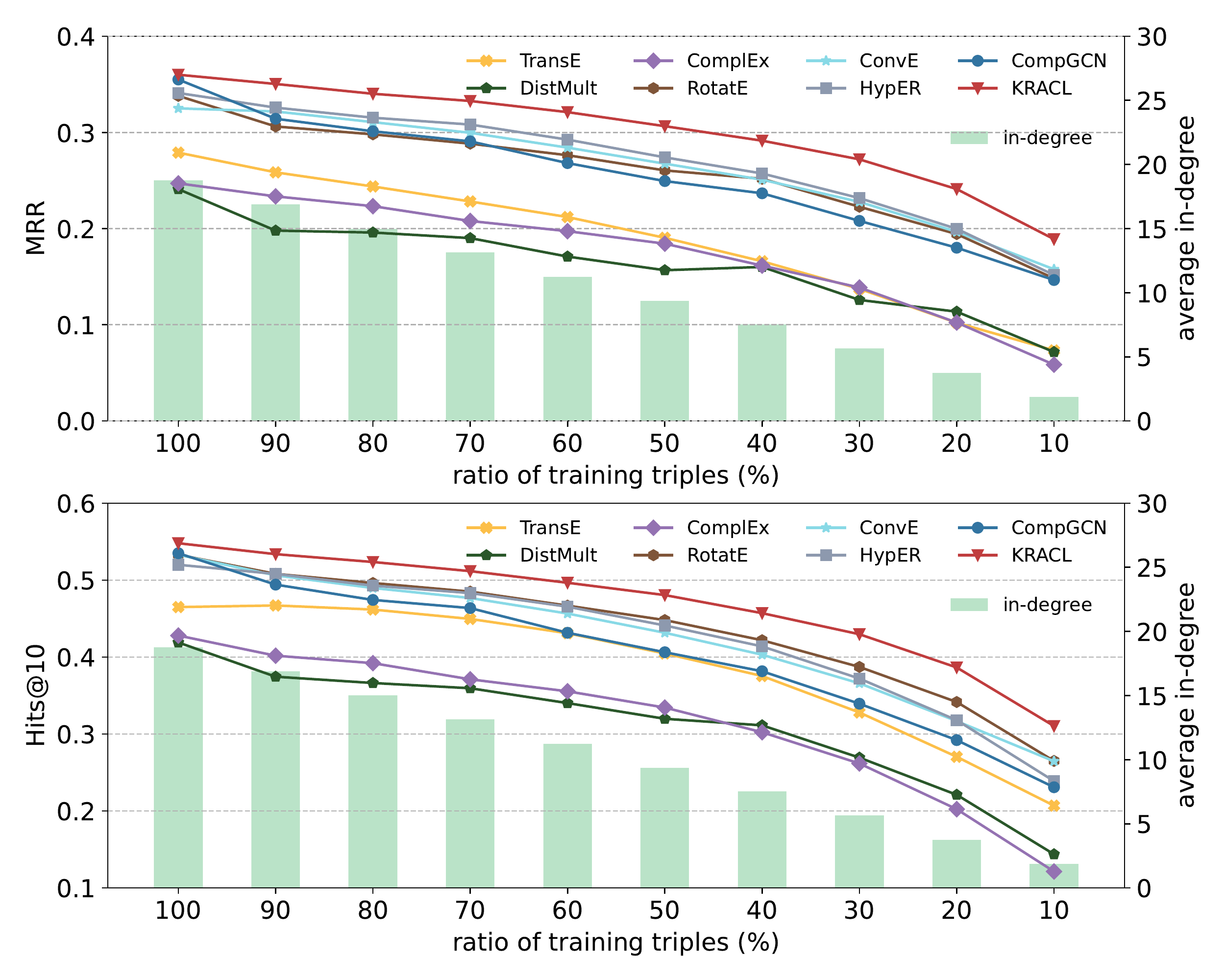}
    \caption{Link prediction performance on sparse knowledge graph of KRACL and competitive peer models on the FB15k-237 datasets.}
    \label{fig:sparsity}
\end{figure}
\subsection{Main Results (RQ1)}
Table \ref{tab:lp_results1} and \ref{tab:lp_results2} show the link prediction performance on the test set on standard benchmarks including FB15k-237, WN18RR, NELL-995, and Kinship\footnote{Please see main results on UMLS dataset in Appendix.}. From the experimental results, we observe that: 1) on sparse knowledge graphs, \emph{i.e.}, WN18RR and NELL-995, KRACL outperforms all other baseline models on most of the metrics. Particularly, MRR is improved from 0.481 and 0.534 in CompGCN to 0.527 and 0.563, about $9.6\%$ and $5.4\%$ relative performance improvement. The MRR of our KRACL also improves upon that of DisenKGAT by a margin of $4.2\%$ and $2.9\%$, respectively. 2) on dense knowledge graphs, \emph{i.e.}, FB15k-237 and Kinship, KRACL also achieves competitive results compared to baseline models, with significant improvement on the Kinship dataset. For the FB15k-237 dataset, we speculate that its abundant N-N relations limit our KRACL's performance. Overall, these results show the effectiveness of the proposed KRACL for the task of predicting missing links in knowledge graphs and its superior performance on both sparse and dense knowledge graphs.

\subsection{Knowledge Sparsity Study (RQ1)}
\label{sec:sparsity}
To verify KRACL's sensitivity against sparsity, we randomly remove triples from the training set of FB15k-237 and evaluate the models on the full test set. Figure \ref{fig:sparsity} shows the MRR  and Hits@10 of 7 competitive models including TransE, DistMult, ComplEx, RotatE, ConvE, HypER, CompGCN, and our proposed KRACL. Performance of all models universally decreases as the training set diminishes. However, the results show that KRACL consistently outperforms all baseline models, and as the corruption ratio increases, the improvement of KRACL against baseline models increases as well. Overall, these experiment results indicate our models' superior robustness against sparsity across a variety of baseline models. 

\begin{table}[t]
    \centering
    \caption{Link prediction performance categorized by different entity in-degree on the FB15k-237 dataset. The best score is in \textbf{bold} and the second best score is \underline{underlined}.}
    \resizebox{1\linewidth}{!}{
    \begin{tabular}{l| c c| c c| c c| c c}
         \toprule[1.5pt] 
         \multirow{2}{*}{\textbf{In-degree}} & \multicolumn{2}{c|}{\textbf{RotatE}} &\multicolumn{2}{c|}{\textbf{ConvE}} & \multicolumn{2}{c|}{\textbf{CompGCN}} &  \multicolumn{2}{c}{\textbf{KRACL}} \\
         & MRR & H@10 & MRR & H@10 & MRR & H@10  & MRR & H@10 \\
        \midrule[0.75pt]
        $[0,10]$ & .178 & .309 & .186 & .338 & \underline{.198} & \underline{.348} & \textbf{.232} & \textbf{.394}\\
        $[10,20]$ & .149 & .294 & .154 & \underline{.299} & \underline{.156} & .296 & \textbf{.181} & \textbf{.335}\\
        $[20, 30]$ & .194 & .381 & \underline{.199} & \underline{.386} & .198 & .370 &\textbf{.218} & \textbf{.405}\\
        $[30, 40]$ &.282 & \underline{.497} & \underline{.287} & .485 & .280 & .476 &\textbf{.307} & \textbf{.501}\\
        $[40, 50]$ & .294 & \underline{.547} & .297 & .516 & \underline{.298} & .520 &\textbf{.328} & \textbf{.552}\\
        $[50, 100]$ &.399 & \underline{.681} & \underline{.403} & .675 & .400 & .663 & \textbf{.434} & \textbf{.702}\\
        $[100, \max]$ &.691 & .929 & \underline{.714} & \textbf{.936} & .674 & .905 & \textbf{.716} & \underline{.932}\\
        \bottomrule[1.5pt]
    \end{tabular}
    }
    
    \label{tab:degree}
\end{table}

\begin{table*}[t]
    \centering
    \caption{Knowledge graph completion performance by relation category on FB15k-237 dataset for TransE, DistMult, ConvE, CompGCN, and proposed KRACL. Following \citet{transh}, the relations are categorized into one-to-one (1-1), one-to-many (1-N), many-to-one (N-1), and many-to-many (N-N).}
    \begin{tabular}{c c|c c|c c|c c|c c|c c}
         \toprule[1.5pt]
         &  & \multicolumn{2}{c|}{\textbf{TransE}} & \multicolumn{2}{c|}{\textbf{DistMult}}& \multicolumn{2}{c|}{\textbf{ConvE}} & \multicolumn{2}{c|}{\textbf{CompGCN}}& \multicolumn{2}{c}{\textbf{KRACL}} \\
         & & MRR & H@10 & MRR & H@10 & MRR & H@10 & MRR & H@10& MRR & H@10\\ 
         \midrule[0.75pt]
         \multirow{4}{*}{\rotatebox{90}{\textbf{Head}}}&1-1 & \underline{.484} & .593& .255& .307& .374 & .505 & .457 & \underline{.604} & \textbf{.500} & \textbf{.609}\\
         &1-N & .080& .152& .038&.071 & .091 & .170 & \underline{.112} & \underline{.190} & \textbf{.118} & \textbf{.215}\\
         &N-1 &.329 & .589& .322&.558 &.444 & .644& \underline{.471} & \underline{.656} & \textbf{.485} & \textbf{.675}\\
         &N-N & .219& .436& .131&.255 & .261 & .459 & \underline{.275} & \underline{.474} & \textbf{.276} & \textbf{.481}\\
         \midrule[0.75pt]
         \multirow{4}{*}{\rotatebox{90}{\textbf{Tail}}}&1-1& \underline{.476} & .588& .257&.312 & .366 & .510 & .453 & \underline{.589} & \textbf{.515} & \textbf{.635}\\
         &1-N &.536 & .846& .575&.750 & .762 & .878 & \underline{.779} &\underline{.885} & \textbf{.796} & \textbf{.894}\\ 
         &N-1& .060& .118& .032&.067 & .069 & .150 & \underline{.076} & \underline{.151} & \textbf{.093} & \textbf{.180}\\
         &N-N & .287&.553 & .184&.376 & .375 & .603 & \textbf{.395}& \underline{.616}& \underline{.394} & \textbf{.620}\\
         \bottomrule[1.5pt]
    \end{tabular}
    \label{tab:cat}
\end{table*}

\subsection{Entity In-degree Analysis (RQ2)}
\label{sec:indegree}
Since the sparsity in KGs will lead to entities with low in-degree and thus lack information to conduct link prediction, we follow \citet{sacn} and analyze link prediction performance on entities with different in-degree. We choose FB15k-237 dataset as our object due to its abundant relation types and dense graph structure. 
As shown in Table \ref{tab:degree}, we present Hits@10 and MRR metrics on 7 sets of entities within different in-degree scopes and compare the performance of KRACL with TransE, DistMult, ConvE, and CompGCN.
Firstly, for entities with low in-degree, GNN-based models such as KRACL and CompGCN outperform ConvE and RotatE, because they get extra information by aggregating neighboring entities. However, we find that simply aggregating neighbors equally through a single operator is not enough. By projecting context triples to different representation spaces, varying the importance of every entity's neighborhood, and introducing more feedback with KCL loss, KRACL achieves significant improvement over all baselines for entities with in-degree $[0, 100]$. For entities with higher in-degree, \emph{i.e.} $[100, \max]$, the performance of KRACL is close to ConvE and RotatE, while the performance of CompGCN is the worst, because entity embedding is substantially smoothed by too much neighboring information \cite{drgi}. To sum up, these results 
show the strong capability of KRACL to predict sparse entities and it is also effective for dense entities.

\begin{figure}[t]
    \centering
    \includegraphics[width=0.86\linewidth]{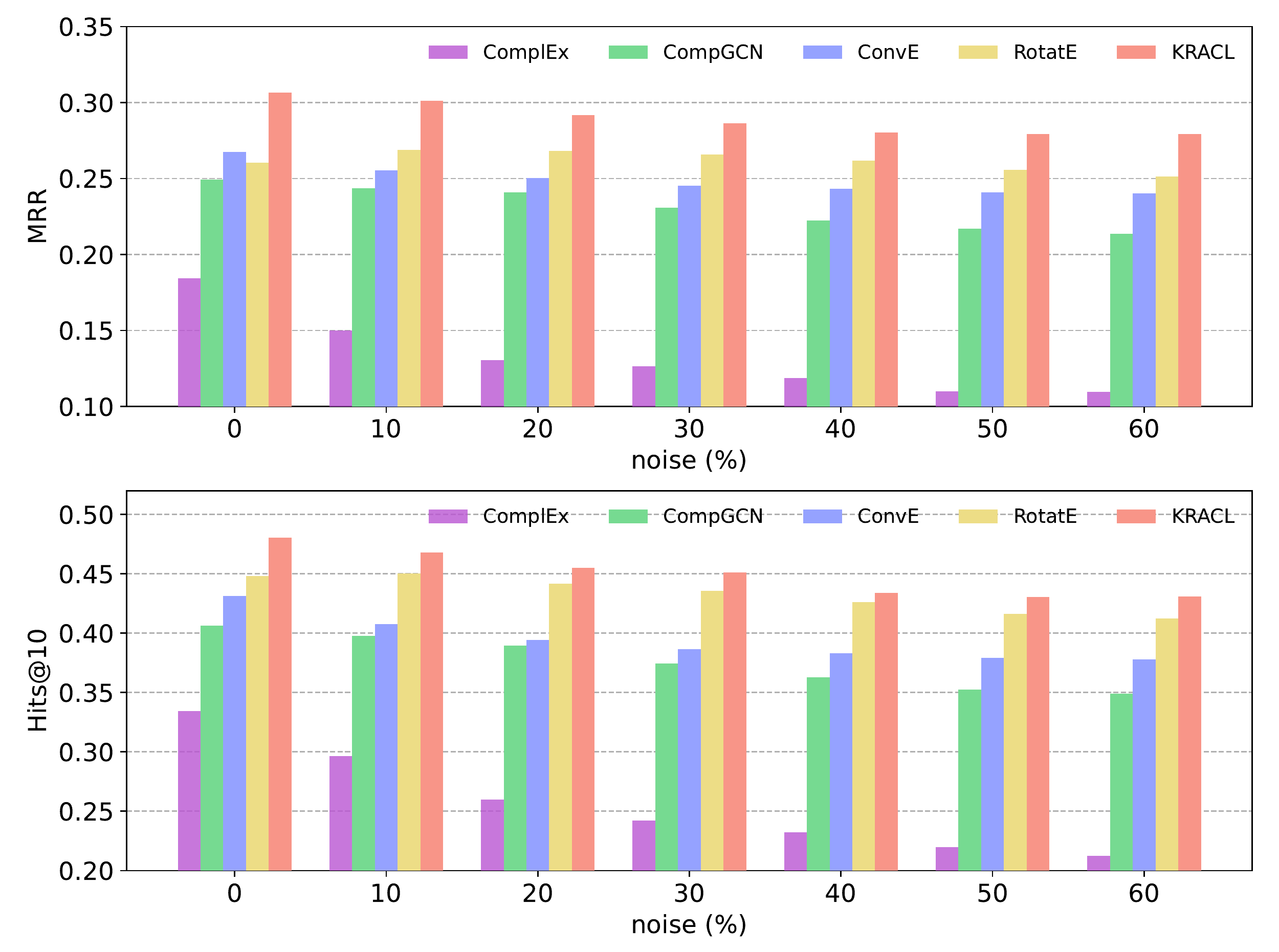}
    \caption{Knowledge graph completion performance on noisy knowledge graph of KRACL and some baseline models on the FB15k-237 dataset.}
    \label{fig:noise}
\end{figure}

\subsection{Performance by Relation Category (RQ2)}
\label{sec:relationcat}
In this part, we follow \citet{transh} and further investigate the performance of KRACL in different relation categories (shown in Table \ref{tab:cat}). We report MRR and Hits@10 of KRACL and compare with TransE, DistMult, ConvE, and CompGCN. We can see that KRACL almost outperforms all baselines for all relation types. Furthermore, it is demonstrated that KRACL achieves significant improvement on 1-1, 1-N, and N-1 relations while the prediction performance on N-N relations is close to CompGCN. We speculate that KRACL is good at learning the relative simple relations and predicting the N-N relation is still challenging to KRACL, which limits model's performance on FB15k-237 dataset. We leave the research of a more expressive scheme to model complex N-N relations as future work.

\begin{table*}[t]
    \centering
    \caption{Knowledge graph completion performance on FB15k-237 dataset. Following \citet{compgcn}, X+M (Y) denotes that M is the GNN backbone to obtain entity and relation embeddings and X is the scoring function or projection head in this work, Y denotes the fusion operator between entity and relation embeddings. The best scores across all settings are highlighted by $\boxed{\cdot}$.}
    \begin{tabular}{l c c c c c c c c}
         \toprule[1.5pt] 
         \textbf{Dec./Proj. (=X) $\rightarrow$} & \multicolumn{2}{c}{\textbf{TransE}} & \multicolumn{2}{c}{\textbf{DistMult}} & \multicolumn{2}{c}{\textbf{RotatE}}& \multicolumn{2}{c}{\textbf{ConvE}} \\
         \cmidrule(r){2-3} \cmidrule(r){4-5} \cmidrule(r){6-7} \cmidrule(r){8-9}  
        \textbf{Methods $\downarrow$} & MRR & H@10 & MRR  & H@10 & MRR & H@10 & MRR & H@10 \\
        \midrule[0.75pt]
        X & .279 & .441 & .241  & .419 &.338 &.533 &.325 & .501 \\
        X+R-GCN & .281 & .469 & .324  & .499 &.295 & .457& .342 & .525 \\
        X+W-GCN & .264 & .444 & .324  & .504 &.272 & .430&.244 & .525 \\
        \hline
        X+CompGCN (Sub) & .335 & .514 & .336 & .513 &.290 & .453& .352 & .530\\
        X+CompGCN (Mult) & .337 & .515 & \textbf{.338} & \textbf{.518} & .296 & .456 &.353 & .532 \\
        X+CompGCN (Rot) & .271 & .447 & .289 &.448 &.296 &.461 &.325 &.506  \\
        X+CompGCN (Corr) & .336 & .518 & .335 & .514 & .294& .459& .355 & .535\\
        \hline
        X+KRAT (Sub) & .334 & .519 & .333 & .512 & .332 & .512 & .355 & .541  \\
        X+KRAT (Mult)& .332 & .513 & .331 & .510 & .334 & .511 & .356 & .540 \\
        X+KRAT (Rot)& .332  & .512 & .331 & .508 & .334 & .513  &  .351 & .538 \\
        X+KRAT (Corr) & .333 & .518 & .334 & .512&.332 & .509 & .353 & .538\\
        X+KRAT (All operators) & \textbf{.340} & \textbf{.524} & \textbf{.338} & .517 & \textbf{.339} & \textbf{.522} & \boxed{\textbf{.360}} & \boxed{\textbf{.548}}\\
        \bottomrule[1.5pt]
    \end{tabular}
    \label{tab:compgcn}
\end{table*}

\subsection{Combination of Different GNN Encoder and Projection Head (RQ3)}
\label{combine}
Borrowing from CompGCN, we evaluate the effect of different GNN methods combined with different knowledge projection heads such as TransE, DistMult, RotatE, and ConvE. The results are shown in Table \ref{tab:compgcn}. We evaluate KRAT on four fusion operators taken from \citet{transe}, \citet{distmult}, \citet{rotate}, \citet{hole}, and with all operators simultaneously.


From experimental results in Table \ref{tab:compgcn}, we have the following observations. First, by utilizing graph neural networks (GNNs), the model can further incorporate graph structure and context information in the knowledge graph and boost model's performance. The lack of fusing relation and entity embeddings leads to poor performance of R-GCN and W-GCN, while CompGCN and KRACL integrate relation and entity context and outperform other baselines. Second, KRAT with all fusion operators outperform all the simple counterparts, which highlights the importance of learning a more power message function in knowledge graphs.
Third, KRACL obtains an average of 4.5\%, 2.8\%, 13.7\%, and 2.5\% relative improvement on MRR compared with CompGCN, which indicates the strong robustness of KRACL across multi-categories knowledge projection heads. We can also see that KRACL significantly outperforms other baseline encoders when combined with RotatE. It reveals the strong robustness and adaptation of the proposed KRACL framework.

\subsection{Robustness against Noisy Triples (RQ3)}
Beyond sparsity, facts generated by knowledge extraction approaches can also be unreliable, \emph{e.g.}, NELL facts have a precision ranging from 0.75-0.85 for confident extractions and 0.35-0.45 across the broader set of extractions \cite{nellsystem}. In this section, we randomly add unreliable triples in the sparse version of FB15k-237 to test the models' robustness against noisy triples. Figure \ref{fig:noise} shows how the MRR and Hits@10 suffer as noises increase. We observe that KRACL consistently outperforms all the baseline models, and its performance shows a lower level of volatility, highlighting its strong robustness against noisy triples.

\begin{table}[t]
    \centering
     \caption{Results of ablation study of the proposed KRACL on the WN18RR and NELL-995 dataset. $BCELoss$ denotes replacing the KCL loss with binary cross entropy loss.}
    \begin{tabular}{l|c c c c c}
         \toprule[1.5pt]
        \multirow{2}{*}{\textbf{Model}} &\multicolumn{2}{c}{\textbf{WN18RR}}&\multicolumn{2}{c}{\textbf{NELL-995}}\\
          & MRR & H@3 & MRR & H@3\\
         \midrule[0.75pt]
        w/o KRAT & .509 &.522 & .543 & .589 \\
        w/o attention & .504 &.521 & .543 & .583\\
        w/o res.& .518 & .532 & .551 & .593 \\
        w/o $\mathcal{L}_{CL}$ & .502 & .514 & .496 & .541\\
        w/o $\mathcal{L}_{CE}$& .495& .531 & .542 & .586\\
        $BCELoss$ & .469 & .478 & .507 & .547 \\
         \hline
        \textbf{KRACL}& \textbf{.527} & \textbf{.547} & \textbf{.563} & \textbf{.602}\\        
        \bottomrule[1.5pt]
    \end{tabular}
   
    \label{tab:ablation}
\end{table}

\subsection{Ablation Study}
\label{sec:ablation}
As KRACL outperforms various baselines across all selected benchmark datasets, we investigate the impact of each module in KRACL to verify their effectiveness. More specifically, we perform ablation studies on the proposed KRAT and its attention mechanism, residual connection, and test the effectiveness of proposed KCL and its two components on WN18RR and NELL-995 datasets, as is shown in Table \ref{tab:ablation}. First, it is illustrated that full KRACL model outperforms 6 ablated models, which proves the effectiveness of our design choice. Second, we observe a significant performance drop when replacing the proposed KCL loss with binary cross entropy loss, which is probably resulted from the poor generalization performance of cross entropy loss when training with limited labels \cite{celosscao, celossliu}.

\section{Conclusion}
In this paper, we present KRACL model to alleviate the widespread sparsity problem for knowledge graph completion. First, KRACL maps context triples to different representation spaces to fully exploit their semantic information and aggregate the messages with the attention mechanism to distinguish their importance. Second, we propose a knowledge contrastive loss to introduce more negative samples, hence more feedback is provided to sparse entities.
Our KRACL effectively improves prediction performance on sparse entities in KGs. Extensive experiments on standard benchmark FB15k-237, WN18RR, NELL-995, Kinship, and UMLS show that KRACL improves consistently over competitive baseline models, especially on WN18RR and NELL-995 with large numbers of low in-degree entities. 

\section*{Acknowledgment}
This work was supported by the National Key Research and Development Program of China (No. 2020AAA0108800), National Nature Science Foundation of China (No. 62192781, No. 62272374, No. 62202367, No. 62250009, No. 62137002), Innovative Research Group of the National Natural Science Foundation of China (61721002), Innovation Research Team of Ministry of Education (IRT\_17R86), Project of China Knowledge Center for Engineering Science and Technology, and Project of Chinese academy of engineering ``The Online and Offline Mixed Educational Service System for `The Belt and Road' Training in MOOC China''. We would like to express our gratitude for the support of K. C. Wong Education Foundation. We also appreciate the reviewers and chairs for their constructive and insightful feedback. Lastly, we would like to thank all LUD lab members for fostering a collaborative research environment.

\bibliographystyle{ACM-Reference-Format}
\bibliography{software}

\newpage
\appendix


\section{Dataset Details}
\begin{itemize}
\item \textbf{FB15k-237} \cite{fb15k237} is a subset of FB15k \cite{transe}, which contains knowledge base describing facts about the real world and is extracted from FreeBase \cite{freebase}. Different from FB15k, it removes all the reverse relations to prevent test data leakage.
\item\textbf{WN18RR} \cite{conve} is a subset of the WordNet \cite{wordnet} containing lecxical relation between words. Similar to FB15k-237, WN18RR also removes the reverse relations to avoid test data leakage.
\item\textbf{NELL-995} \cite{nell995} is a subset of the 995-th iteration of NELL system. From Table \ref{tab:benchmark} we can see that it is much sparser than other datasets.
\item\textbf{Kinship} \cite{kinship} contains a set of triples that explains the kinship relationships among members of the Alyawarra tribe from Central Australia. It is an integral part of aboriginal across Australia with regard to marriages between aboriginal people.
\end{itemize}

\section{Relation Category Details}
Following \citet{transh}, for each relation $r$, we compute the average number of tails per head and the average number of head per tail, denoted as $tphr$ and $hptr$, respectively. If $tphr<1.5$ and $hptr<1.5$, $r$ is treated as one-to-one (1-1); if $tphr<1.5$ and $hptr\geq 1.5$, $r$ is treated as many-to-one (N-1); if $tphr\geq 1.5$ and $hptr<1.5$, $r$ is treated as one-to-many (1-N); if $tphr\geq 1.5$ and $hptr\geq 1.5$, $r$ is treated as a many-to-many (N-N).

\begin{figure}[t]
    \centering
    \includegraphics[width=1\linewidth]{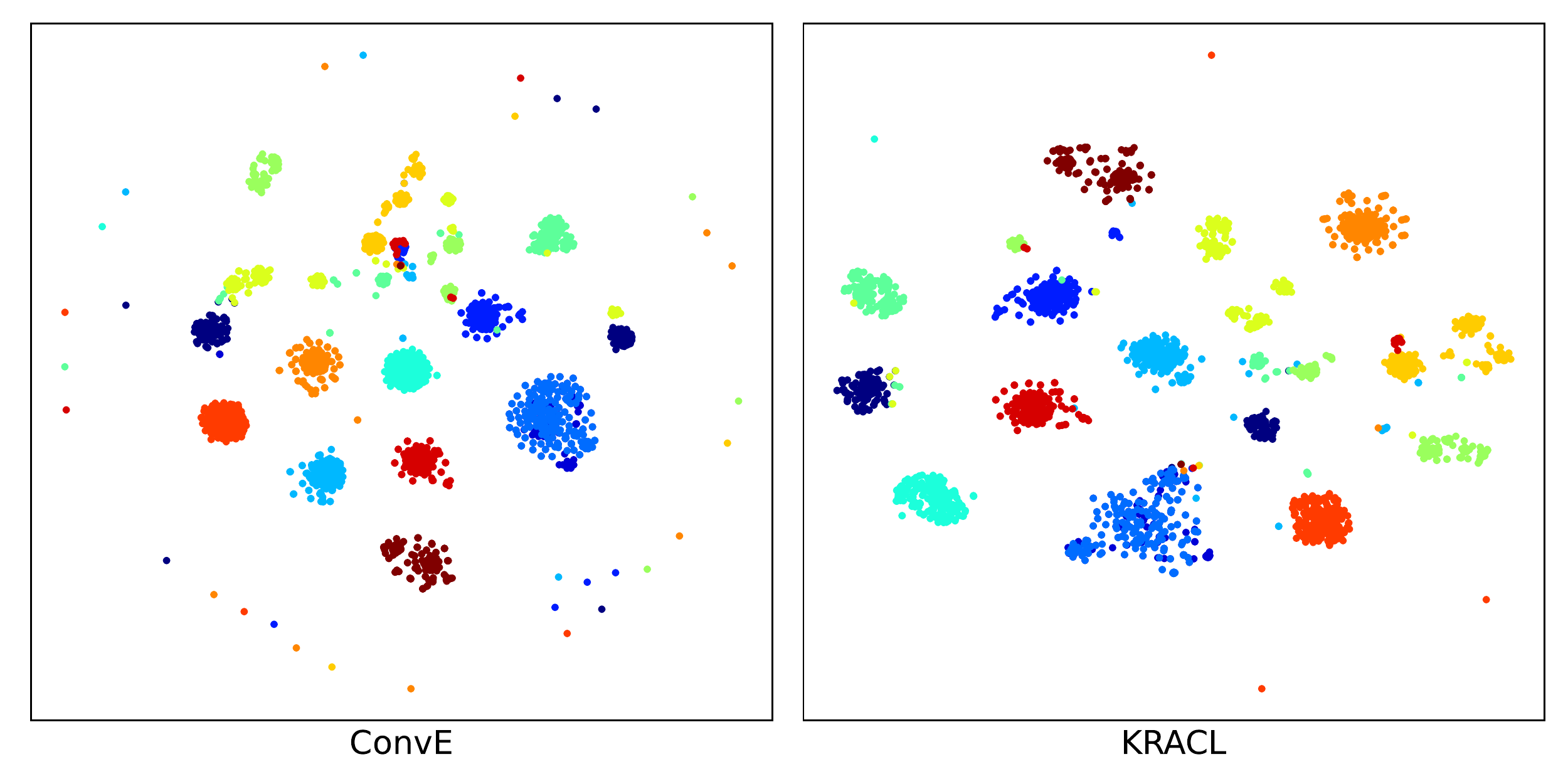}
    \caption{Visualization of object entity embeddings in ConvE and KRACL with T-SNE. }
    \label{fig:tsne}
\end{figure}

\begin{table}[t]
    \centering
    \caption{Number of parameters in the KRACL model and GPU hours for training on selected datasets.}
    \begin{tabular}{l|c c c}
        \toprule[1.5pt] 
        \textbf{Dataset} & \textbf{Parameters} & \textbf{GPU hours}\\
        \midrule[0.75pt] 
         \textbf{FB15k-237} & 13.3M & 9.5\\
         \textbf{WN18RR} & 18.6M & 4.5\\
         \textbf{NELL-995} & 25.9M & 10.5 \\ 
         \textbf{Kinship} & 10.4M & 0.7\\
         \textbf{UMLS} & 10.4M & 0.5\\
        \bottomrule[1.5pt] 
    \end{tabular}
    
    \label{tab:parameter}
\end{table}

\section{Visualization of Entity Representations}
To examine the quality of learned representations, we visualize the entity embeddings. Given a link prediction task $(s,r,?)$, we select queries $(s,r,?)$ that have the same answers and visualize their predictions with T-SNE \cite{tsne}. As is shown in Figure \ref{fig:tsne}, our model shows higher level of collocation for entities, which indicates that our KRACL framework learns high-quality representations for entities and relations.

\section{Implementation Details}

 We implement our KRACL model in PyTorch \cite{pytorch}, PyTorch Lightning \cite{pytorchlightning}, and Pytorch Geometric \cite{torchgeometric} library on an RTX 3090 GPU with 24GB memory. Following \citet{compgcn}, each triple $(s,r,o)$ is augmented with a flipped triple $(o, r^{-1}, s)$. We present our hyperparameter settings in Table \ref{tab:hyper} to facilitate reproducibility. We also use OpenKE \cite{openke} and PyKEEN \cite{pykeen} to reproduce the baselines.

For the main results shown in Table \ref{tab:lp_results1} and \ref{tab:lp_results2}, we adjust the hyperparameters based on the performance on the validation set and report the best results on the test set. For other experiments, we present the performance of a single run. 

\begin{table}[t]
    \centering
    \caption{Hyperparameter settings of KRACL across various benchmark datasets. We find our hyperparameter settings robust across all datasets and all hyperparameters are chosen by the performance on the validation set.}
    \resizebox{1\linewidth}{!}{
    \begin{tabular}{l|c c c c c}
        \toprule[1.5pt] 
        \textbf{Hyperarameter} & \textbf{FB15k-237} & \textbf{WN18RR} & \textbf{NELL-995} & \textbf{Kinship} & \textbf{UMLS} \\
        \midrule[0.75pt] 
        Entity dim $d_e$ &200 & 200 & 200 & 200 & 200 \\
        Relation dim $d_r$ &200 & 200 & 200 & 200 & 200 \\
        Batch size & 2048 & 2048 & 2048 & 1024 & 1024\\
        Learning rate & $\rm 10^{-3}$ & $\rm 10^{-3}$ & $\rm 10^{-3}$ & $\rm 3\times10^{-4}$ & $\rm 5\times10^{-4}$\\
        Epochs & 1500 & 1000 & 1000 & 1000 & 1000\\
        GNN layers & 1 &2 & 2& 2 & 2\\
        Encoder dropout & 0.1 & 0.2 & 0.2 & 0.2 & 0.2\\
        Temperature $\tau$ &0.07 & 0.07 & 0.07 & 0.1 & 0.1\\
        Optimizer & AdamW & AdamW & AdamW & AdamW & AdamW\\
       \bottomrule[1.5pt] 
    \end{tabular}
    }
    
    \label{tab:hyper}
\end{table}

\begin{table}[t]
    \centering
    \caption{Link prediction performance of KRACL and several baseline models on the UMLS dataset. The best score is in \textbf{bold} and the second best score is \underline{underlined}.}
    \begin{tabular}{l|c c c c}
         \toprule[1.5pt] 
         \multirow{2}{*}{\textbf{Model}} & \multicolumn{4}{c}{\textbf{UMLS}} \\
         & MRR $\uparrow$& MR $\downarrow$& H@10 $\uparrow$& H@1 $\uparrow$\\
         \midrule[0.75pt]
          TransE & .615 & 3.6 & .945 & .391\\
          DistMult & .164 & 18.8 & .403  & .061\\
          ComplEx & .844 & 2.47 & .967  & \underline{.765}\\
          RotatE & .822 & 2.1 & .969 & .703\\
          ConvE & .836 & 3.2 & .946  &.764\\
          ConvKB & .782 & \underline{1.61} & .986  & .593\\
          SACN  & \underline{.856} & 1.7 & .985 & .764 \\
          R-GCN & .481 & 7.8 &.835 & .318 \\
          KBAT & .818 & 1.855 & \underline{.987} & .711\\
          \hline
          \textbf{KRACL} & \textbf{.904} & \textbf{1.38} & \textbf{.995} & \textbf{.831}\\
         \bottomrule[1.5pt]
    \end{tabular}
    
    \label{tab:umls}
\end{table}

\section{Fusion Operator Details}
\label{app:op}
 \begin{itemize}
\item \textbf{Rotation}: $\phi(\boldsymbol{h}_s, \boldsymbol{h_r}) = \boldsymbol{h}_s \circ \boldsymbol{h}_r$\\
 For each dimention $i$, $e[2i]$ and $e[2i+1]$ are corresponding real and imaginary components. Given the subject embedding $e_s$ and relation transform embedding $\theta_r$, the rotation projection is formulated as
 \begin{equation}
 \begin{aligned}
&\left[
  \begin{array}{c} 
    (\boldsymbol{h}_s \circ \boldsymbol{h}_r)[2i]\\ 
    (\boldsymbol{h}_s \circ \boldsymbol{h}_r)[2i+1] \\ 
  \end{array}
\right]=\\
&\left[
  \begin{array}{c c} 
    \cos \boldsymbol{h}_r(i) & -\sin \boldsymbol{h}_r(i)\\ 
    \sin \boldsymbol{h}_r(i) & \cos \boldsymbol{h}_r(i) \\ 
  \end{array}
\right]
\left[
  \begin{array}{c} 
    \boldsymbol{h}_s[2i]\\ 
    \boldsymbol{h}_s[2i+1] \\ 
  \end{array}
\right],
\end{aligned}
 \end{equation}
 where $\theta_r$ is learnable parameter corresponding to relation type $r$, $\hat{h_o}$ denotes the projected object embedding after rotation.
 
\item \textbf{Circular-correlation}: $\phi(\boldsymbol{h}_s, \boldsymbol{h_r}) = \boldsymbol{h}_s \star \boldsymbol{h}_r$

Taken from \citet{hole}, the circular-correlation operator is formulated as
\begin{equation}
(h_s\star h_r)[k] = \sum_{i=0}^{d-1} h_{s}[i] \cdot h_r[(k+i)\; mod\; d],
\end{equation}
where $d$ is the dimension of entity and relation embeddings, $mod$ denotes the modulo operation. The circular-correlation operator can discriminate the direction of relation because of its non-commutative property.
\end{itemize}

\end{document}